\documentclass[letterpaper, 10 pt, conference]{ieeeconf}  

\usepackage{cite}

\usepackage{multicol}
\usepackage{notoccite}
\usepackage{graphicx}
\usepackage{subcaption}
\usepackage{amsmath}
\usepackage{amssymb}
\usepackage[linesnumbered,ruled,vlined]{algorithm2e}
\usepackage{booktabs} 
\usepackage{multirow} 
\usepackage{siunitx}
\usepackage{float}
\usepackage[table]{xcolor}
\usepackage{tablefootnote}

\usepackage[hidelinks, colorlinks=false]{hyperref}
\usepackage{cleveref}
\usepackage{tikz}
\usepackage{atbegshi}
\usepackage{url}

\IEEEoverridecommandlockouts                            

\title{\LARGE \bf
Accelerated Multi-Modal Motion Planning Using Context-Conditioned Diffusion Models}

\author{Edward Sandra$^{1,3}$ \and Lander Vanroye$^{1,3}$ \and Dries Dirckx$^{1,3}$ \and Ruben Cartuyvels$^{2}$ \and Jan Swevers$^{1,3}$ \and Wilm Decr{\'e}$^{1,3}$%
\thanks{$^{1}$Department of Mechanical Engineering, KU Leuven, 3001 Heverlee, Belgium}%
\thanks{$^{2}$Department of Computer Science, KU Leuven, 3001 Heverlee, Belgium}%
\thanks{$^{3}$Flanders Make@KU Leuven, Belgium}%
\thanks{Correspondence to: Edward Sandra \texttt{edward.sandra@kuleuven.be}}%
\thanks{The implementation is publicly available at \url{https://meco-group.github.io/campd}.}
}

\begin{document}

\maketitle
\thispagestyle{empty}
\pagestyle{empty}

\begin{abstract}
      Classical methods in robot motion planning, such as sampling-based and optimization-based methods, often struggle with scalability towards higher-dimensional state spaces and complex environments. Diffusion models, known for their capability to learn complex, high-dimensional and multi-modal data distributions, provide a promising alternative when applied to motion planning problems and have already shown interesting results. However, most of the current approaches train their model for a single environment, limiting their generalization to environments not seen during training. The techniques that do train a model for multiple environments rely on a specific camera to provide the model with the necessary environmental information and therefore always require that sensor.
      To effectively adapt to diverse scenarios without the need for retraining, this research proposes Context-Aware Motion Planning Diffusion (CAMPD).
      CAMPD leverages a classifier-free denoising probabilistic diffusion model, conditioned on sensor-agnostic contextual information.
      An attention mechanism, integrated in the well-known U-Net architecture, conditions the model on an arbitrary number of contextual parameters. CAMPD is evaluated on a 7-DoF robot manipulator and benchmarked against state-of-the-art approaches on real-world tasks, showing its ability to generalize to unseen environments and generate high-quality, multi-modal trajectories, at a fraction of the time required by existing methods.
\end{abstract} 

\section{Introduction}

Robot motion planning is a fundamental problem in robotics, involving the task of finding a collision-free trajectory for a robot between a start and goal configuration. Solving this problem can be challenging, particularly in high-dimensional and complex environments. High-dimensional robot motion planning problems naturally arise from robots with many degrees of freedom (DoF), as well as when multiple robots share the same environment and a motion plan must be planned for all robots simultaneously. Complex environments emerge because robots often need to operate in cluttered spaces. Moreover, motion planners are frequently required to consider constraints beyond avoiding collisions, such as dynamic constraints to ensure the feasibility of the motion on the robot. Additionally, the motion planner should be highly efficient to respond effectively to disturbances in the environment within a limited time frame during online deployment.
Classical approaches frequently encounter limitations in addressing these challenges.
They can be split up into two families: sampling-based methods and optimization-based methods. Sampling-based methods~\cite{rrt,rrtconnect, prm}, while asymptotically complete, produce non-smooth trajectories and often struggle to scale efficiently as problem dimensions increase. Optimization-based methods~\cite{chomp, trajopt, gpmp2, curobo, fatrop} on the other hand, can produce solutions that adhere to arbitrary constraints and optimize specific objectives, but are prone to becoming trapped in local minimizers and are sensitive to the initializations. 

Recently, learning-based techniques have emerged as a promising alternative to classical motion planning algorithms in robotics, with methods broadly categorized into learning from demonstration on the one hand and reinforcement learning~(RL)~\cite{rl} on the other hand.
Learning from demonstration trains neural networks on expert trajectories, generated either by humans or classical algorithms, and can replace components or entire motion planners.
Standalone approaches~\cite{ETE1, ETE2, plannwithdiff} generate collision-free trajectories directly, while hybrid methods integrate deep learning with classical planners to enhance performance, such as guiding sampling processes~\cite{hybrid-example-cnn} or providing initial guesses for optimization-based planners~\cite{hybrid-cvae,mpd,diffusionseeder}.
In contrast, RL uses trial-and-error interactions with the environment to optimize a reward function, eliminating the need for expert datasets. Policies learned via RL can function as standalone planners or complement existing methods~\cite{surveyRL}.
Learning-based methods hold promise for addressing high-dimensional planning problems, but challenges remain, including limited generalization to unseen environments during training, substantial data requirements for learning from demonstrations, and the complexities of reward shaping and inefficient learning in reinforcement learning.

Several of these promising, emerging methods use diffusion models~\cite{diffusion_origin, ddpm, sbm_origin, survey_diffusion_planning, plannwithdiff,mpd,diffusionseeder}. This generative model family has shown success in a variety of applications, including image generation, video generation and speech generation~\cite{survey_diffusion}. In robotics, diffusion models have been applied to both imitation~\cite{imitation1,imitation2,dallebot} and policy learning~\cite{diffpolicy, diffpolicy2, diffpolicy3, jackson2024policyguideddiffusion, ren2024diffusionpolicypolicyoptimization}.
However, important research challenges remain~\cite{survey_diffusion_planning}, particularly in enhancing the generalization and robustness of these deep learning-based methods. Existing diffusion models are typically not environment-aware, which limits their adaptability to new environments~\cite{diffpolicy,plannwithdiff}. While environment information can be incorporated via gradient guidance~\cite{mpd, edmp}, this approach increases computational cost. Techniques that train models to be environment-aware rely on a camera to provide the necessary environmental information, requiring an additional processing step and increasing the algorithm's complexity~\cite{diffusionseeder}.

This work addresses the generalization challenge by introducing Context-Aware Motion Planning Diffusion (CAMPD), a classifier-free diffusion probabilistic model (DPM) for robot motion planning, capable of rapidly generating multi-modal solutions and adapting to unseen environments and tasks. 
CAMPD is a planning-as-inference approach~\cite{planning-as-inference}, where collision-free, near-optimal plans are generated by sampling from the model's learned conditional distribution. An attention mechanism~\cite{attention} is integrated within the U-Net~\cite{unet} architecture to condition the model on encoded environmental parameters. 
After training with occasional context dropout, classifier-free guidance is employed during inference to steer the sampled trajectories towards collision-free regions. The proposed method is evaluated in simulation on a 7-DoF Franka Emika Panda robot in sphere-based and real-world environments, both with varying number of obstacles. Results are compared against a classical planner and state-of-the-art learning-based methods. 

The key contributions of this work are as follows:
\begin{enumerate}
    \item \textbf{Context-aware diffusion-based motion planner:} A diffusion-based motion planner is introduced that integrates sensor-agnostic contextual information, such as obstacle locations, directly into the model.
    \item \textbf{Improved generalization:} Extensive experiments provide evidence that CAMPD achieves significantly better generalization to novel environments compared to state-of-the-art planners.
    \item \textbf{Real-time, executable trajectories:} CAMPD generates high-quality, dynamically feasible trajectories in real time, enabling rapid planning and replanning in dynamic settings without the need for post-processing.
\end{enumerate}

The proposed approach is evaluated on a diverse set of challenging motion planning problems, demonstrating strong generalization to previously unseen environments and effectively capturing the multi-modal nature of locally optimal solutions. CAMPD directly outputs robot-executable trajectories using an efficient post-processing step. Its computational efficiency makes it particularly well-suited for real-time, online applications, enabling faster planning and replanning in dynamic environments at a fraction of the computational cost required by existing methods.


\begin{figure*}
      \centering
      \includegraphics[width=1\textwidth]{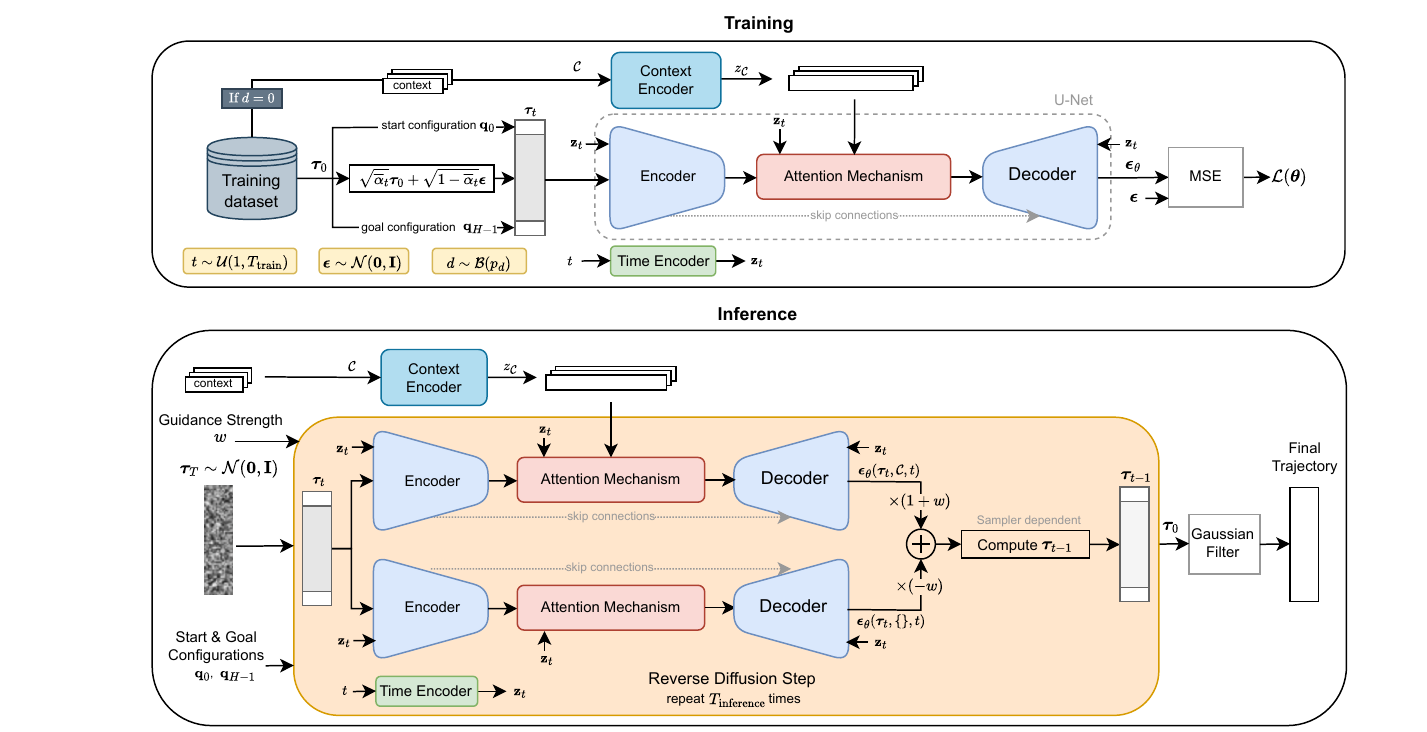}
      \caption{Overview of CAMPD: During training, trajectories are perturbed with Gaussian noise at a random diffusion step, while start and goal states remain fixed for conditioning. Context is encoded and randomly dropped (Bernoulli $p_d$) for conditional/unconditional training. A U-Net with attention predicts the noise using MSE loss. During inference, noise is iteratively denoised via reverse diffusion with classifier-free guidance ($w$), keeping start and goal fixed. Finally, a Gaussian filter is applied to the resulting trajectory to reduce jerk.
 }
      \label{fig:overview}
\end{figure*}

\section{Related Work}
Leveraging denoising diffusion probabilistic models~\cite{diffusion_origin,ddpm} has shown promise in generating multi-modal plans for high-dimensional problems. Janner et al.~\cite{plannwithdiff} introduced a method in which sampling from the model and planning become nearly identical. However, the model needs to be trained per environment, limiting its adaptability to different environments. MPD~\cite{mpd} and EDMP~\cite{edmp} integrate the environment into their methods by exploiting cost functions to guide the diffusion process, thereby generating collision-free trajectories. While their method allows for generalization to unseen environments during inference, the learned network lacks direct knowledge of the environments and only relies on these cost functions to achieve collision avoidance. Furthermore, hand-crafted cost functions require tuning and can lead to increased computation time. SceneDiffuser~\cite{scenediffuser} and DiffusionSeeder~\cite{diffusionseeder} condition their model on environmental information. SceneDiffuser utilizes point clouds as environmental input, while DiffusionSeeder employs depth images encoded by a Vision Transformer~\cite{vit}. Moreover, DiffusionSeeder leverages the trajectories sampled from the diffusion model as priors for cuRobo~\cite{curobo}, an optimization-based planner, effectively blending learning-based and optimization-based approaches while maintaining low computation times. While these methods yield promising results, they rely on environmental representations such as point clouds or camera data, which may introduce unnecessary complexity when a simpler description of the environment (e.g. precise object dimensions and locations) is available. Furthermore, this abstraction of the environment representation allows flexibility in how environmental information is obtained. Decision Diffuser~\cite{decisiondiffuser} similarly conditions on contextual information, but it does not support an arbitrary number of contextual elements, such as a variable number of obstacles, during inference.
In contrast to these methods, CAMPD leverages a classifier-free DPM, conditioned on sensor-agnostic contextual information, supporting an arbitrary number of contextual elements.

\section{Methodology}
\label{sec:method}
This work studies the robot motion planning problem, where the goal is to rapidly generate a fast, smooth, collision-free trajectory to bring a robot manipulator from an initial configuration to a goal configuration.

\subsection{Terminology} 
In this work, a trajectory $\boldsymbol{\tau} \triangleq (\mathbf{q}_0, \ldots, \mathbf{q}_{H-1}) \in \mathbb{R}^{d_q \times H}$ is defined as a sequence of robot configurations in discrete-time with horizon $H$ and configuration space dimension $d_q$. The context $\mathcal{C}$ of a trajectory $\boldsymbol{\tau}$ is defined as a structured collection of relevant environmental information and other parameters that provide useful insights for the task. The context $\mathcal{C} = \{ \mathbf{c}_{k,l} \}$ can be categorized into $K$ distinct types (e.g. obstacle types), where each type $k=0, \dots, K-1$ can have multiple instances, each represented as a vector $\mathbf{c}_{k,l}$ with specific parameters (e.g. context type $1$ are $L$ spheres with parameters $\mathbf{c}_{1,l} = \left[ x_l,y_l,z_l,r_l \right]$ for $l=0, \dots, L-1$). The start and goal configurations $\mathbf{q}_{0}$ and $\mathbf{q}_{H-1}$ are not considered part of the context.

\subsection{Neural Network Architecture}
As in MPD~\cite{mpd}, CAMPD employs planning-as-inference via a diffusion probabilistic model (DPM)~\cite{diffusion_origin, ddpm}. A DPM consists of two processes: a forward and reverse diffusion process. The noise $\boldsymbol{\epsilon}$ added in the forward process is learned using the mean squared error (MSE) loss:
\begin{equation}
      \label{eq:loss}
      \mathcal{L}(\boldsymbol{\theta}) = \mathbb{E}_{t, \boldsymbol{\epsilon}, \boldsymbol{\tau}_0, \mathcal{C}} \left[ \| \boldsymbol{\epsilon} - \boldsymbol{\epsilon_\theta} (\boldsymbol{\tau}_t, \mathcal{C},t) \|^2           \right],
  \end{equation}
  with $\boldsymbol{\theta}$ the network parameters and $\boldsymbol{\epsilon_\theta}$ the learned estimate of $\boldsymbol{\epsilon}$.
The estimated mean $\boldsymbol{\mu}_\theta$ of the reverse process is then given by~\cite{ddpm}
\begin{equation}
      \label{eq:2bckgnd-mu}
      \boldsymbol{\mu}_\theta(\boldsymbol{\tau}_t,\mathcal{C},t) = \frac{1}{\sqrt{\alpha_t}}\left(\boldsymbol{\tau}_t - \frac{\beta_t}{\sqrt{1-\overline{\alpha}_t}} \boldsymbol{\epsilon_\theta}(\boldsymbol{\tau}_t,\mathcal{C},t) \right).
  \end{equation}
  
\begin{figure}
      \centering
          \includegraphics[width=0.5\textwidth]{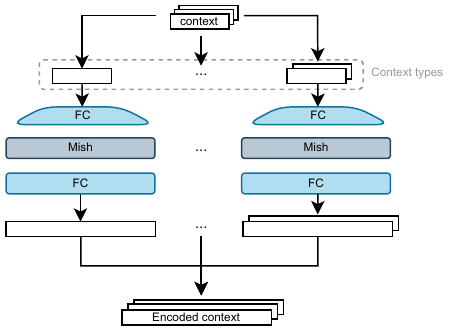}
          \caption{Context Encoder. Each instance $\mathbf{c}_{k,l}$ of context type $k$ is processed by a dedicated multi-layer perceptron (MLP)}
      \label{fig:context_encoder}
\end{figure}

  In CAMPD, the diffusion model estimates the added noise $\boldsymbol{\epsilon}$ in both conditional and unconditional modes, such that classifier-free guidance~\cite{cfg} can be employed:
  \begin{equation}
        \label{eq:2cfg}
        \boldsymbol{\epsilon_\theta}'(\boldsymbol{\tau}_t,\mathcal{C},t) = (1 + w) \boldsymbol{\epsilon_\theta}(\boldsymbol{\tau}_t, \mathcal{C}, t) - w \boldsymbol{\epsilon_\theta}(\boldsymbol{\tau}_t, \{ \}, t)
    \end{equation}
  with $w$ the guidance strength.
  The network architecture consists of a time encoder, context encoder and a U-Net, as illustrated in \Cref{fig:overview}. 
\paragraph{Time Encoder}
The time encoder applies sinusoidal positional encoding~\cite{attention} followed by a single-hidden-layer multi-layer perceptron (MLP), yielding a latent representation $\mathbf{z}_t = \text{MLP}_t(\text{PE}(t)) \in \mathbb{R}^{d_z}$ to condition the U-Net on the diffusion step.
\paragraph{Context Encoder}
The context encoder maps the context information $\mathcal{C}$ to a set of latent representations, denoted as $z_\mathcal{C} = \{\mathbf{z}_{\mathbf{c}_{k,l}} \mid \mathbf{z}_{\mathbf{c}_{k,l}} \in \mathbb{R}^{d_z}\}$. Each instance $\mathbf{c}_{k,l}$ of context type $k$ is processed through a single-hidden-layer $\text{MLP}_k$, yielding the latent representation $\mathbf{z}_{\mathbf{c}_{k,l}} = \text{MLP}_k(\mathbf{c}_{k,l})$. A schematic overview of the context encoder is provided in \Cref{fig:context_encoder}. These latent representations are subsequently utilized in the attention mechanism of the U-Net. 

\begin{figure}
      \centering
          \includegraphics[width=0.43\textwidth]{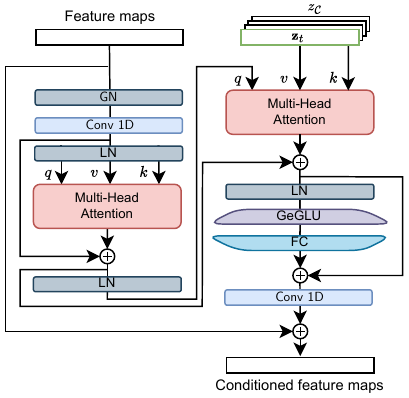}
          \caption{Attention Mechanism. The encoded context vectors $\mathbf{z}{\mathbf{c}_{k,l}}$ and the latent time step representation $\mathbf{z}_t$ are fused into the U-Net through a residual architecture of self- and cross-attention layers.}
      \label{fig:attention_unet}
\end{figure}

\paragraph{U-Net}
The U-Net, denoted as $\boldsymbol{\epsilon}_{\boldsymbol{\theta}, \textbf{U}}(\boldsymbol{\tau}_t, z_\mathcal{C}, \mathbf{z}_t)$, estimates the added noise $\boldsymbol{\epsilon}$, given the noisy trajectory $\boldsymbol{\tau}_t$ and the latent time and context representations $\mathbf{z}_t$ and $\mathbf{z}_{\mathbf{c}_{k,l}}$. The convolutional encoder and decoder networks are similar to the one used by \cite{plannwithdiff} and are conditioned on the latent time step representation $\mathbf{z}_t$. Between the encoder and decoder, an attention mechanism is integrated to condition the model on the context information. This allows the model to incorporate an arbitrary number of context instances $\mathbf{c}_{k,l}$ (e.g. environments with varying numbers of obstacles), rather than conditioning the model on a fixed context encoding. 
The attention mechanism is detailed in \Cref{fig:attention_unet}. The attention mechanism is inspired by the Transformer architecture~\cite{attention}. It consists of a multi-head self-attention mechanism and a multi-head attention mechanism with the latent representations $\mathbf{z}_t$ and $\mathbf{z}_{\mathbf{c}_{k,l}}$ as both keys and values, followed by a feed-forward network.

\subsection{Training}
Given a noisy trajectory $\boldsymbol{\tau}_t$, a time step $t$ and context $\mathcal{C}$, the model learns to predict the added noise $\boldsymbol{\epsilon}$ via the mean squared error (MSE) loss given in \Cref{eq:loss}.
To enable classifier-free guidance during inference, the model is trained in both conditional and unconditional modes by setting the context information to an empty set following a Bernoulli distribution $\mathcal{B}(p_d)$ with probability $p_d$. To make the model conditioned on the start and goal configurations, these configurations are not perturbed during training.
The training algorithm is detailed in \Cref{alg:training} and illustrated in \Cref{fig:overview}.

\begin{algorithm}[t]
\caption{Training}\label{alg:training}
\KwIn{Training set $\mathcal{D}$, Diffusion model $\boldsymbol{\epsilon_\theta}$, learning rate $\alpha$, noise schedule parameters $\{\overline{\alpha}_t\}$, probability of unconditional training $p_d$}
\While{not converged}{
\tcp{Sample batch of trajectories and related contexts}
$\{ (\boldsymbol{\tau}_0^i, \mathcal{C}^i) \}_{i=1}^N \sim \mathcal{D}$ \;
\tcp{Randomly discard contexts to train conditionally or unconditionally} 
$b^i \sim \mathcal{B}(p_d), \; \forall i$ \;
$\mathcal{C}^i \leftarrow \{ \} $ \textbf{if }$b^i = 1, \; \forall i$ \;
$t^i \sim \mathcal{U}(1, T_{\text{train}}), \; \boldsymbol{\epsilon}^i \sim \mathcal{N}(\mathbf{0}, \mathbf{I}), \; \forall i$ \;
\tcp{Evaluate noised trajectories}
$\boldsymbol{\tau}_{t}^i \leftarrow \sqrt{\overline{\alpha}_{t_i}} \boldsymbol{\tau}_0^i + \sqrt{1 - \overline{\alpha}_{t_i}} \boldsymbol{\epsilon}^i, \; \forall i$ \;
\tcp{Set start and goal configurations}
$\boldsymbol{\tau}_{t,0}^i \leftarrow \boldsymbol{\tau}_{0,0}^i, \; \boldsymbol{\tau}_{t,H-1}^i \leftarrow \boldsymbol{\tau}_{0,H-1}^i \; \forall i$ \;
$\boldsymbol{\epsilon}_0^i \leftarrow 0, \; \boldsymbol{\epsilon}_{H-1}^i \leftarrow 0 \;  \forall i$ \;
\tcp{Compute batch loss}
$\mathcal{L}(\boldsymbol{\theta}) \leftarrow \frac{1}{N} \sum_{i=1}^N || \boldsymbol{\epsilon}^i - \boldsymbol{\epsilon_\theta} (\boldsymbol{\tau}_t^i, \mathcal{C}^i, t^i) ||^2$ \;
\tcp{Update model parameters}
$\boldsymbol{\theta} \leftarrow \boldsymbol{\theta} - \alpha \nabla_{\boldsymbol{\theta}} \mathcal{L}(\boldsymbol{\theta})$ \;
}
\end{algorithm}

\begin{algorithm}[t]
            \caption{Inference}\label{alg:inference}
            \KwIn{Diffusion model $\boldsymbol{\epsilon_\theta}$, start configuration $\mathbf{q}_0$, goal configuration $\mathbf{q}_{H-1}$, context $\mathcal{C}$, guidance strength $w$, sampling algorithm $f_\text{denoise}$ }
            \tcp{Initialization}
            $\boldsymbol{\tau}_{T_{\text{inf}}} \sim \mathcal{N}(\mathbf{0}, \mathbf{I})$ \;
            \tcp{Set start and goal configurations}
                $\boldsymbol{\tau}_{T_{\text{inf}},0} \leftarrow \mathbf{q}_0, \; \boldsymbol{\tau}_{T,H-1} \leftarrow \mathbf{q}_{H-1}$ \;
            \tcp{Encode context}
            $z_\mathcal{C} \leftarrow \{ \text{MLP}_k(\mathbf{c}_{k,l}) \mid \mathbf{c}_{k,l} \in \mathcal{C} \}$ \;
            \For{$t = T_{\text{inf}}, \dots, 1$}{
                  \tcp{Encode time step}
                  $\mathbf{z}_t \leftarrow \text{MLP}_t(\text{PE}(t))$ \;
                  \tcp{Classifier-free guidance}
                  $\boldsymbol{\epsilon}_t' \leftarrow (1 + w) \boldsymbol{\epsilon}_{\boldsymbol{\theta}, \textbf{U}}(\boldsymbol{\tau}_t, z_\mathcal{C}, \mathbf{z}_t) - w \boldsymbol{\epsilon}_{\boldsymbol{\theta}, \textbf{U}}(\boldsymbol{\tau}_t, \{ \}, \mathbf{z}_t)$ \;
                  \tcp{Denoise via the sampling algorithm}
                  $\boldsymbol{\tau}_{t-1} \leftarrow f_\text{denoise}(\boldsymbol{\tau}_t, \boldsymbol{\epsilon}_t', t)$ \;
      
                \tcp{Set start and goal configurations}
                $\boldsymbol{\tau}_{t,0} \leftarrow \mathbf{q}_0, \; \boldsymbol{\tau}_{t,H-1} \leftarrow \mathbf{q}_{H-1}$ \;
          
            }
      
            \tcp{Apply Gaussian filter}
            $\boldsymbol{\tau}_0 \leftarrow \text{GaussianFilter}(\boldsymbol{\tau}_0)$ \;
            \KwOut{Trajectory $\boldsymbol{\tau}_0$}
\end{algorithm}


\subsection{Inference}
Trajectories are generated by iteratively denoising white Gaussian noise through a reverse diffusion process. The form of this process depends on the chosen sampler: it can be a discrete-time stochastic process (e.g., DDPM~\cite{ddpm}), a deterministic non-Markovian process (e.g., DDIM~\cite{ddim}), or the numerical solution of a differential equation using high-order solvers (e.g., DPM-Solver++~\cite{dpmsolverplusplus}).
To strengthen the effect of contextual information during inference, classifier-free guidance~\cite{cfg} is applied (\Cref{eq:2cfg}).
To ensure the trajectory begins and ends at the desired configurations, the first and last states of the noisy sequence \( \boldsymbol{\tau}_t \) are fixed to the start and goal states, respectively. Finally, a Gaussian filter is applied to the resulting trajectory to reduce jerk. The complete inference procedure is described in \Cref{alg:inference} and illustrated in \Cref{fig:overview}.

  \begin{table*}
      \centering
      \caption{Evaluation on sphere-based environments with $N_{\text{batch}} = 100$. Results are reported as mean ± standard deviation. Time (T) denotes the duration to generate one batch of trajectories. Success (S) indicates whether at least one feasible trajectory exists in the batch. Feasible Trajectories Rate (FTR) is the percentage of collision-free trajectories with all joints within limits. Best Smoothness Difference (BSD) reflects the relative smoothness gap between the best CAMPD trajectory and that of the baseline. Variance (Var) quantifies the sum of distance variances across corresponding joint configurations over the trajectory dimension.}
      \label{tab:results}
      \begin{tabular}{@{}lllccccc@{}}
      \toprule
      \textbf{Experiment} & \textbf{Method} &  & \textbf{T (s)} $\downarrow$ & \textbf{S ($\%$)} $\uparrow$ & \textbf{FTR ($\%$)} $\uparrow$ & \textbf{BSD ($\%$)} $\downarrow$ & \textbf{Var} $\uparrow$\\
      \midrule
      \multirow{4}{*}{\begin{tabular}{@{}l}
      Random Spheres
      \end{tabular}} & \multicolumn{2}{l}{RRTC $+$ Fatrop}             & 16.49 $\pm$ 14.70* & 86.6 $\pm$ 34.1 & -- & --   & 0.5 $\pm$ 1.4\\
      & \multirow{4}{*}{\textbf{CAMPD}}  & $w = 1$ & \textbf{0.066 $\pm$ 0.0} & \textbf{97.5 $\pm$ 15.6} & 79.4 $\pm$ 31.0  & \textbf{2.2 $\pm$ 7.0} & 1.7 $\pm$ 3.3 \\
      &            & $w = 1.5$       & 0.066 $\pm$ 0.0 & 97.4 $\pm$ 15.9 & 82.3 $\pm$ 30.4 &  3.9 $\pm$ 10.7& 1.9 $\pm$ 3.7\\
      &            & $w = 2$       & 0.066 $\pm$ 0.0 & 97.0 $\pm$ 17.1 & \textbf{82.8 $\pm$ 30.7} &  5.0 $\pm$ 12.4 & 2.2 $\pm$ 4.5\\
      &            & $w = 5$         & 0.066 $\pm$ 0.0 & 97.0 $\pm$ 17.1 & 67.9 $\pm$ 31.5 &  9.7 $\pm$ 45.2 & \textbf{3.7 $\pm$ 8.3}\\
      \midrule
      \multirow{2}{*}{\begin{tabular}{@{}l}
            Partially Random Spheres
      \end{tabular}} & \multicolumn{2}{l}{MPD}               & 3.165 $\pm$ 0.8 & 80.7 $\pm$ 39.5 & 59.6 $\pm$ 43.4 &  -- & \textbf{4.8 $\pm$ 6.8}\\
      & \multirow{1}{*}{\textbf{CAMPD}}            & $w = 1.5$         & \textbf{0.066 $\pm$ 0.0} & \textbf{90.7 $\pm$ 29.1} & \textbf{73.3 $\pm$ 37.5} & \textbf{-36.0 $\pm$ 14.5} &1.7 $\pm$ 3.1\\
      \bottomrule
      \addlinespace[0.25mm]
      \multicolumn{1}{l}{\small *Under ideal parallel CPU execution.}
      \end{tabular}
\end{table*}

\begin{table*}[ht]
      \centering
      \small
      \setlength{\tabcolsep}{5pt}
      \caption{Evaluation on the $\text{M}\pi\text{Net}$~\cite{mpinets} test set. Results for DiffusionSeeder are taken from~\cite{diffusionseeder}.}
      \label{tab:realworld}
      \resizebox{0.9\textwidth}{!}{%
      \begin{tabular}{@{}l!{\vrule width 1.5pt}c|cc|cccc!{\vrule width 1.5pt}c|cccc@{}}
          & \multicolumn{7}{c!{\vrule width 1.5pt}}{$N_{\text{batch}}=12$} & \multicolumn{5}{c}{$N_{\text{batch}}=64$} \\
          \cmidrule(lr){2-8} \cmidrule(lr){9-13}
          & \multicolumn{1}{c|}{cuRobo} 
          & \multicolumn{2}{c|}{DiffusionSeeder} 
          & \multicolumn{4}{c!{\vrule width 1.5pt}}{IK cuRobo + \textbf{CAMPD}} 
          & \multicolumn{1}{c|}{cuRobo} 
          & \multicolumn{4}{c}{IK cuRobo + \textbf{CAMPD}} \\
          & & & 
          & \scriptsize DDPM & \multicolumn{3}{|c!{\vrule width 1.5pt}}{\scriptsize DPMSolver++} 
          & & \scriptsize DDPM & \multicolumn{3}{|c}{\scriptsize DPMSolver++} \\
          & & $N_{\text{iters}}=25$ & 50 
          & \multicolumn{1}{c}{$w=1$} & \multicolumn{1}{c}{1} & 2 & 5 
          &  & \multicolumn{1}{c}{$w=1$} & 1 & 2 & 5 \\
          \midrule
          Plan Time (ms) & 26 & 15 & 17 & 7+20 & \textbf{8+5} & 8+5 & 8+5 & 111 & 8+52 & \textbf{8+7} & 8+7 & 8+7 \\
          \midrule
          Success Rate & \textbf{92.0\%} & 85.1\% & 85.8\% & 90.7\% & 86.9\% &  87.6\% & 78.9\% & 93.8\% & \textbf{98.3\%} &96.2\% & \textbf{97.2\%} &93.3\% \\
          \midrule
          Jerk ($\text{rad/}\text{s}^3$) & 36.5 & 108.8 & 103.6 & 26.1 & \textbf{25.9} & 29.1 & 41.6 & 62.2 & \textbf{37.1} & 37.2 & 42.0 & 75.6 \\
          Motion Time (s) & \textbf{1.12} & 1.26 & 1.26 & 1.38 & 1.36 & 1.37 &1.47 & \textbf{1.01} & \textbf{1.13} & 1.14 & 1.14 & 1.22 \\
          Translation Err (mm) & 0.60 & 0.98 & 0.95 & 0.004 & \textbf{0.003} & 0.003 & 0.003 & 0.87 & \textbf{0.003 }& 0.003 & 0.003 & 0.003 \\
          Quaternion Err ($^\circ$) & 0.13 & 0.93 & 1.44 & 0.005 & \textbf{0.004} &0.004  & 0.004 & 0.16 & \textbf{0.004} & 0.004 & 0.004 & 0.004 \\
          \bottomrule
      \end{tabular}
      }
  \end{table*}

\begin{figure}[htpb]
      \begin{subfigure}{.15\textwidth}
            \centering
            \includegraphics[width=\textwidth]{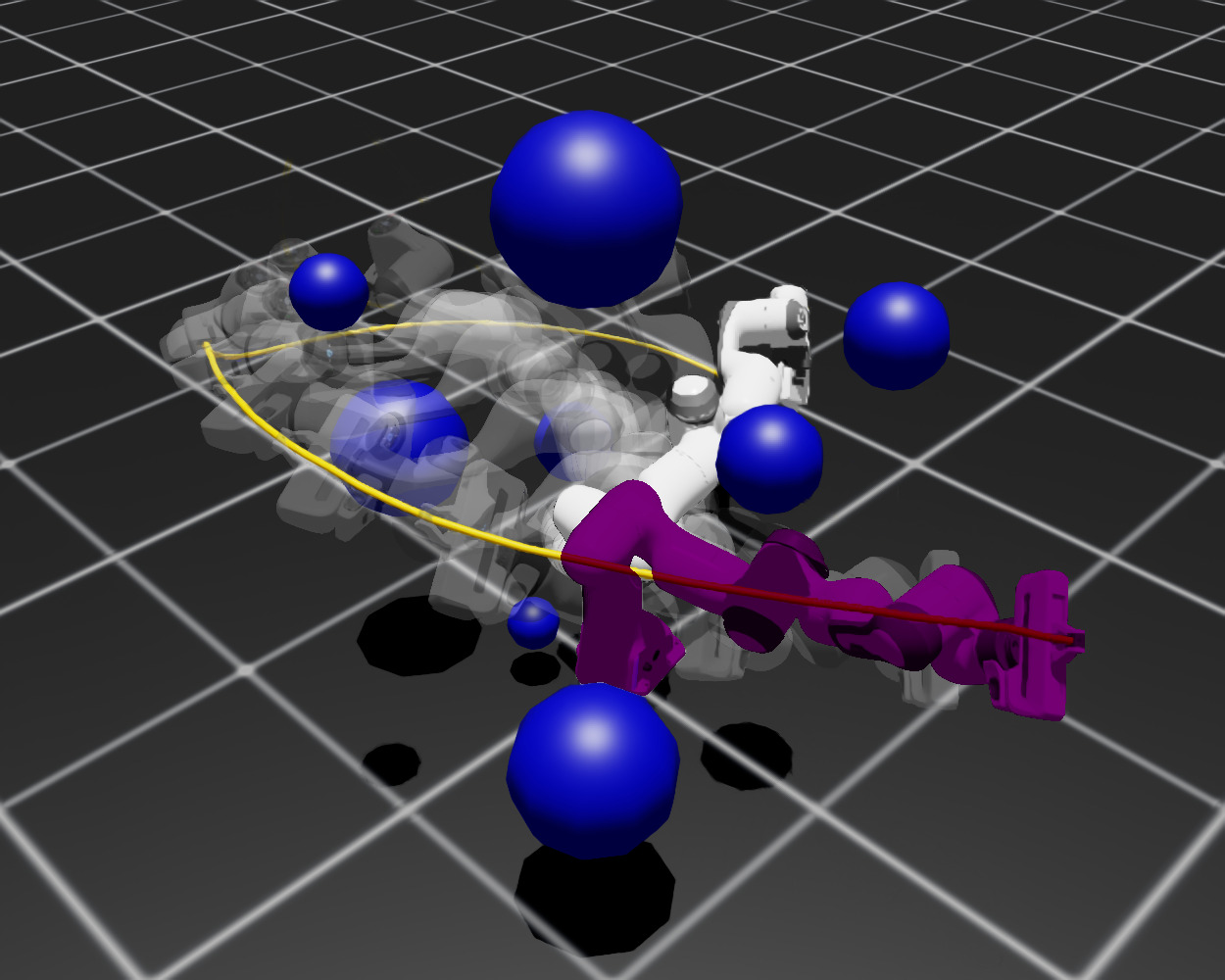}
            \caption{}
            
      \end{subfigure}
      \begin{subfigure}{.15\textwidth}
            \centering
            \includegraphics[width=\textwidth]{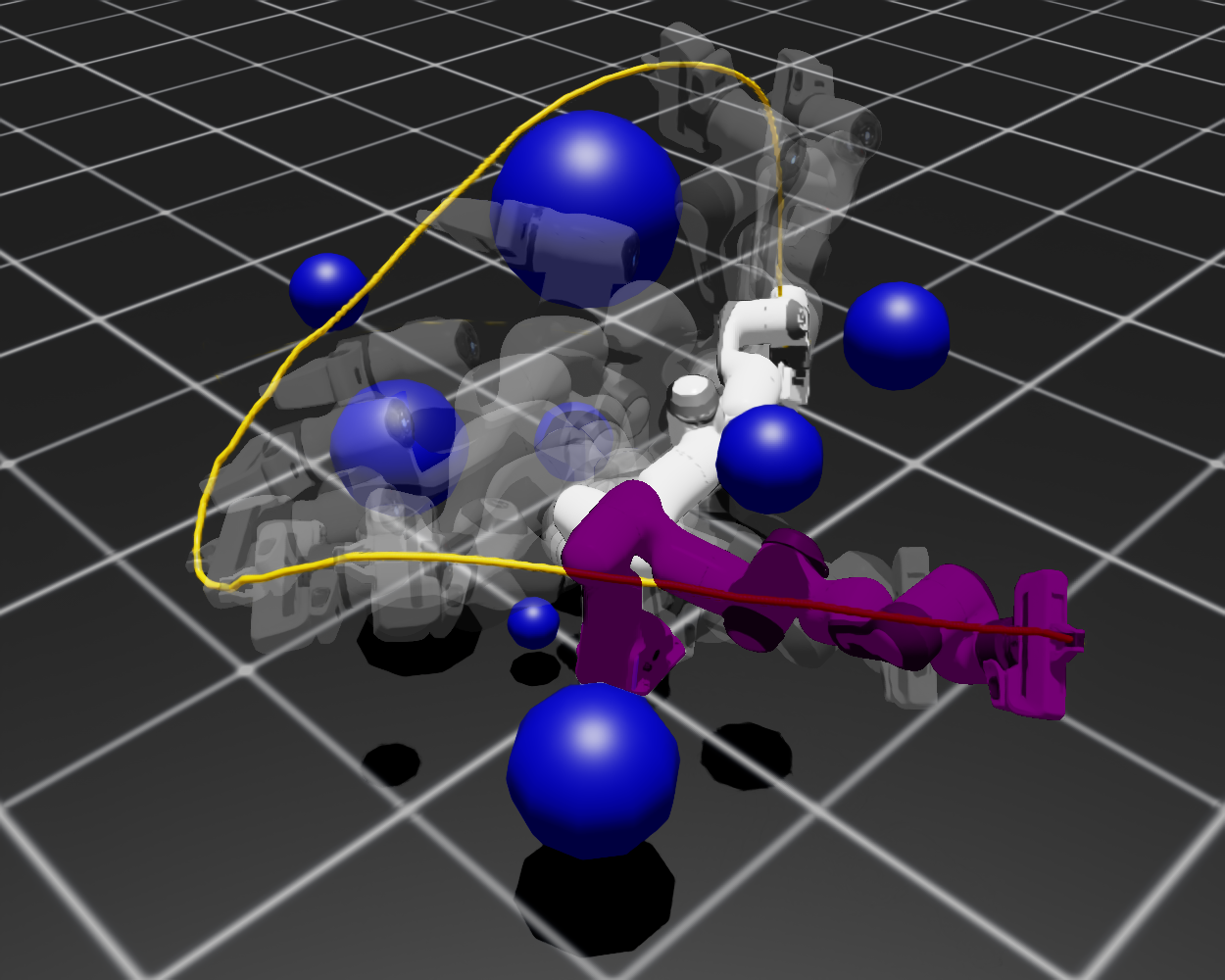}
            \caption{}
            
      \end{subfigure}
      \begin{subfigure}{.15\textwidth}
            \centering
            \includegraphics[width=\textwidth]{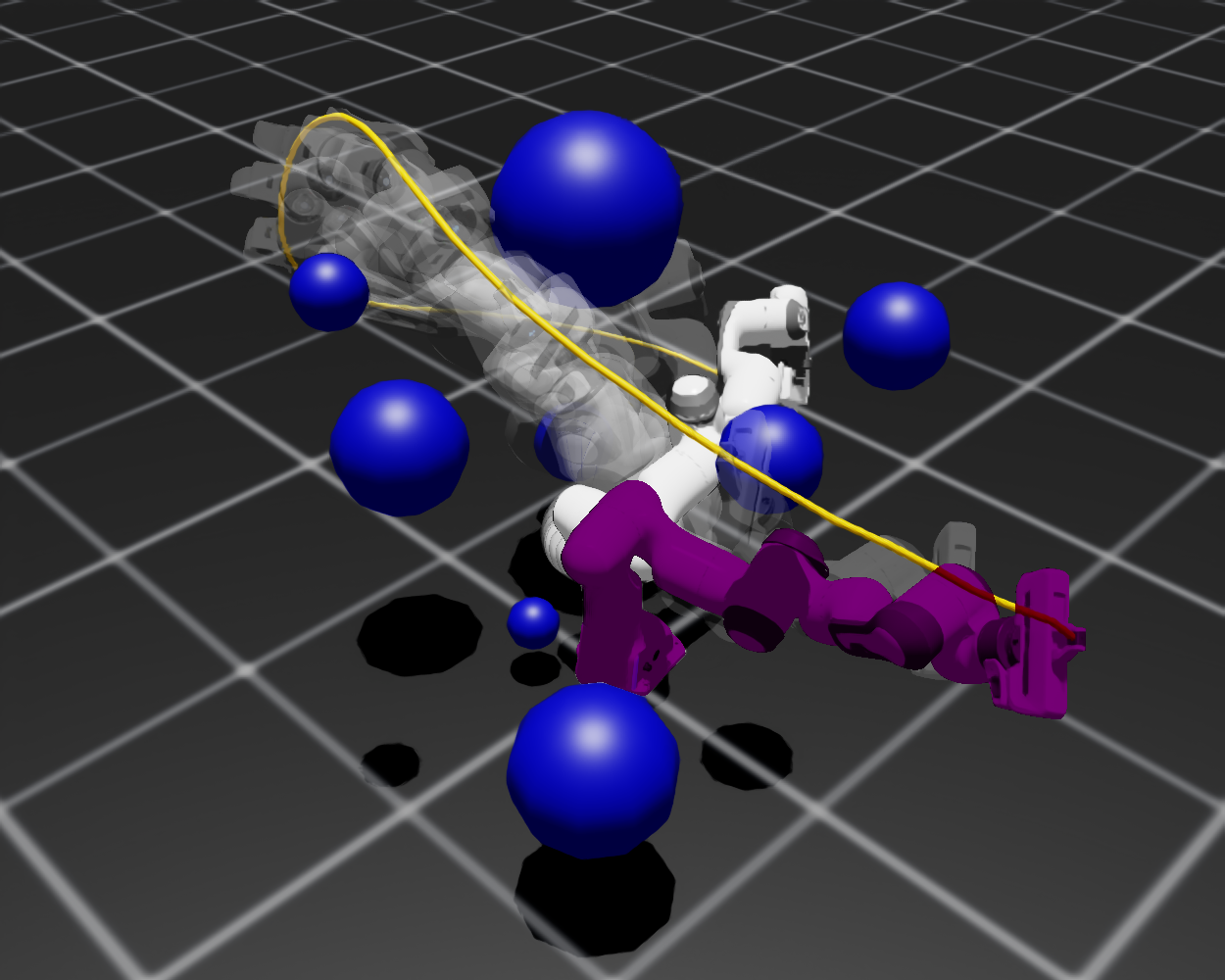}
            \caption{}
  \label{fig:examplec}
      \end{subfigure}
      \caption{Example of multi-modal trajectories generated by CAMPD in an unseen environment with spherical obstacles (blue). The white and purple arms indicate start and goal configurations, respectively. The yellow line shows the end-effector trajectory, with (c) highlighting the best trajectory of a batch of 100 samples.}
      \label{fig:example}
  \end{figure}

  \begin{figure}
      \begin{subfigure}{.15\textwidth}
            \centering
            \includegraphics[width=\textwidth]{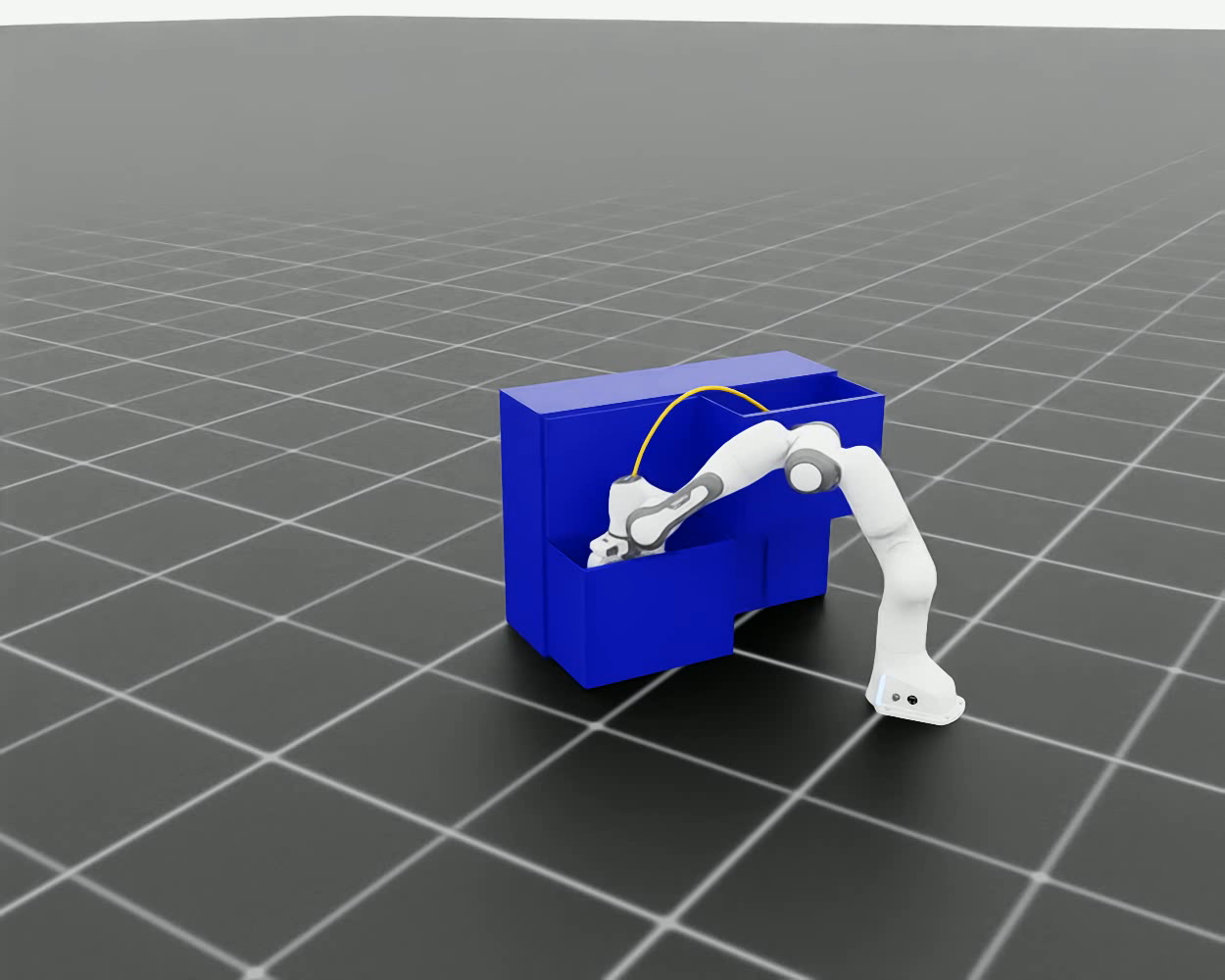}
            
      \end{subfigure}
      \begin{subfigure}{.15\textwidth}
            \centering
            \includegraphics[width=\textwidth]{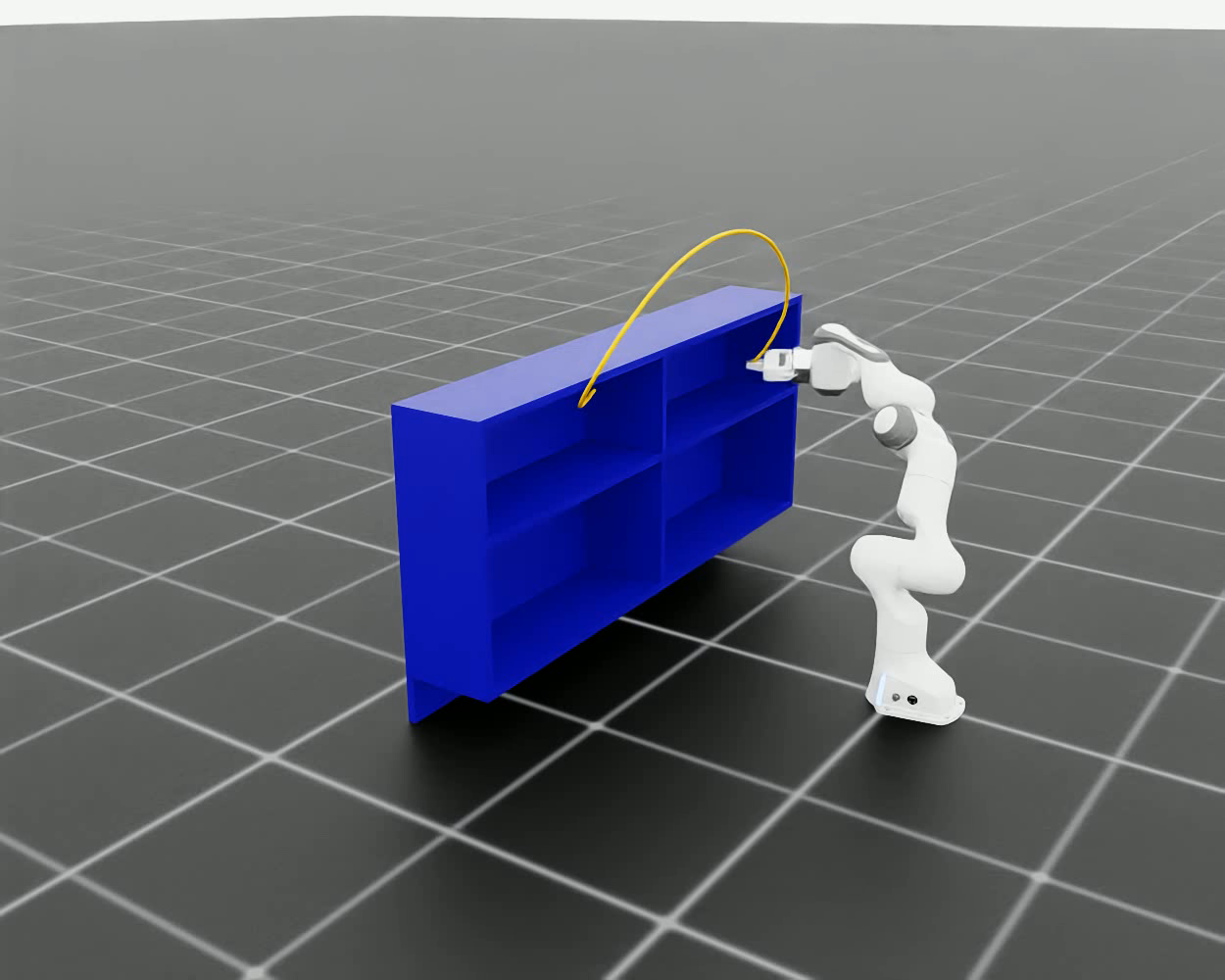}
            
      \end{subfigure}
      \begin{subfigure}{.15\textwidth}
            \centering
            \includegraphics[width=\textwidth]{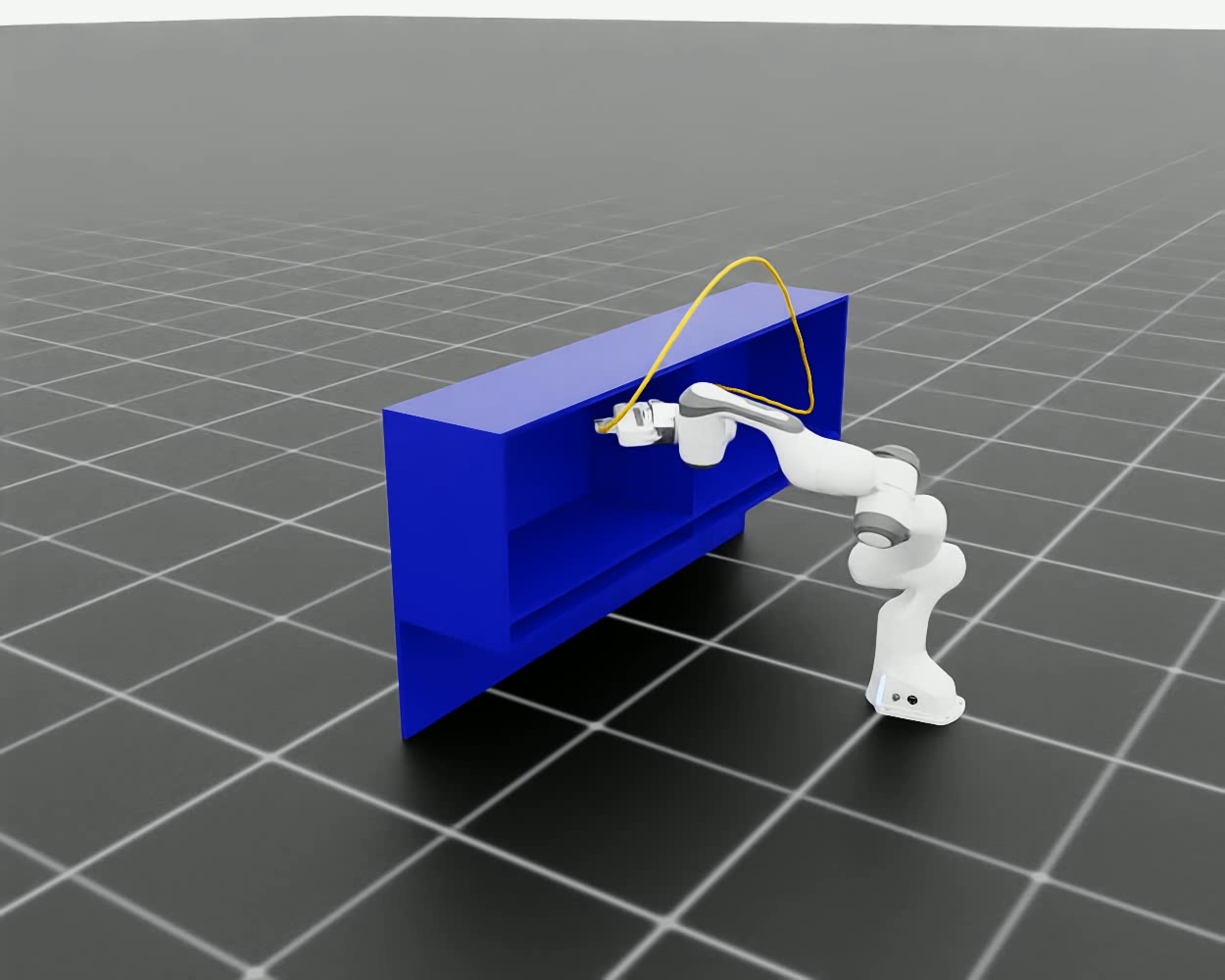}
      \end{subfigure}
      \caption{Feasible trajectories generated by CAMPD for three previously unseen problems from the $\text{M}\pi\text{Net}$~\cite{mpinets} test set.}
      \label{fig:example_mpinets}
  \end{figure}

\section{Experimental Results}
To evaluate CAMPD, experiments were conducted using a 7-DoF Franka Emika Panda robot. CAMPD is compared against a classical motion planner (RRT-Connect~\cite{rrtconnect} + Fatrop~\cite{fatrop}) that combines sampling-based and optimization-based techniques, as well as MPD~\cite{mpd}. Additional evaluations were performed on the test set of the $\text{M}\pi\text{Net}$~\cite{mpinets} dataset, which features simulated real-world robotic tasks, comparing CAMPD to cuRobo~\cite{curobo} and DiffusionSeeder~\cite{diffusionseeder}. The same architecture and set of hyperparameters is used for all experiments, unless otherwise specified. See Appendix~\ref{sec:modelparams} for more details.

\subsection{Evaluation in Simulated Sphere-Based Environments}
The robot is tasked with navigating from a start configuration to a goal configuration through cluttered environments containing a varying number of spheres between one and ten. The randomly sized and placed spheres are spaced such that there is sufficient free space between them for the robot to pass through.
The context consists of sphere instances, each represented by a vector $\mathbf{c}_{\text{spheres},l} = \left[ x_l, y_l, z_l, r_l \right]$ containing the position and radius.
The training set was generated using a hybrid planner of RRT-Connect~\cite{rrtconnect} and Fatrop~\cite{fatrop}, resulting in a total of \num{113469} locally optimal and smooth trajectories across \num{2400} distinct environments. The reverse sampling process was performed using DDPM~\cite{ddpm}.

\Cref{fig:example} shows an example of multi-modal solutions generated by CAMPD in an unseen environment.
The results are summarized in \Cref{tab:results}. On a first test set of 1000 environments not seen during training, CAMPD outperforms the hybrid planner in both success rate and variance. The positive smoothness difference suggests that, on average, the trajectories generated by CAMPD are less smooth than those from the hybrid planner. Nonetheless, there are still cases where CAMPD finds more optimal solutions by identifying modes that the hybrid planner missed. Surprisingly, CAMPD surpasses the method used to generate its own dataset in these metrics. Intuitively, this may occur because the diffusion model learns a global distribution over feasible trajectories from many planning instances, allowing it to sample solutions in regions of the solution space that the hybrid planner may fail to find a feasible trajectory.

In a second test set of 600 unseen environments (\Cref{tab:results}, second row),  involving partially fixed obstacles, CAMPD outperforms MPD in terms of success rate, feasible trajectory rate, and best smoothness difference, indicating that it is more successful in finding feasible and optimal solutions. This is particularly surprising, as part of the environment is already seen during MPD training, while it was not present in the CAMPD dataset. 
However, because MPD's optimization costs balance each other, MPD finds a more diverse set of solutions, but they tend to be less smooth.

In both experiments, the guidance strength $w$ has a significant impact on the performance of CAMPD. A higher guidance strength pushes the trajectories further away from those generated by the unconditional model (a straight line in joint space in this case, as that is the minimal mean squared error if the trajectories in the training set are equally distributed over the robot's joint space). This can result in a higher success rate and feasible trajectory rate. However, a higher guidance strength often leads to less smooth solutions, as indicated by the best smoothness difference.

CAMPD is computationally efficient and deterministic, taking only \SI{0.066}{\second} per batch of 100 trajectories, compared to \SI[separate-uncertainty=true]{16.49(14.73)}{\second} per batch of 100 trajectories for the hybrid planner, whose slowness stems from low RRT-Connect success rates, and \SI[separate-uncertainty=true]{3.165(0.818)}{\second} per batch of 100 trajectories for MPD, whose slowness arises from cost-gradient computations.

              \subsection{Evaluation of Simulated Real-World Tasks}
              To evaluate CAMPD in real-world scenarios, a model was trained to solve tasks from the $\text{M}\pi\text{Net}$ simulation test set~\cite{mpinets}. Each task specifies an initial configuration and a target end-effector pose, and includes realistic scenes with cuboids and cylinders forming furniture-like setups. 
              Since CAMPD requires a goal configuration, the inverse kinematics (IK) solver from cuRobo~\cite{curobo} is used during inference to generate a batch of feasible joint-space configurations from the target pose. The task context consists of cuboids and spheres, each represented by feature vectors: $\mathbf{c}_{\text{cuboid},l} = [x_l, y_l, z_l, w_l, h_l, d_l, q_{w,l}, q_{x,l}, q_{y,l}, q_{z,l}]$ and $\mathbf{c}_{\text{sphere},l} = [x_l, y_l, z_l, r_l, h_l, q_{w,l}, q_{x,l}, q_{y,l}, q_{z,l}]$. Here, $(x_l, y_l, z_l)$ denotes the object position, $(w_l, h_l, d_l)$ the cuboid dimensions, $r_l$ the sphere radius, and $(q_{w,l}, q_{x,l}, q_{y,l}, q_{z,l})$ the orientation quaternion. This context is explicitly provided as part of the task description.
               The training dataset, generated using cuRobo~\cite{curobo}, contains 1 million trajectories across \num{200000} distinct environments, drawn from the same distribution as the $\text{M}\pi\text{Net}$ test set.

               CAMPD is evaluated against two baselines: cuRobo~\cite{curobo} and DiffusionSeeder~\cite{diffusionseeder}. \Cref{tab:realworld} reports results on the $\text{M}\pi\text{Net}$ test set, with baseline metrics for DiffusionSeeder taken from~\cite{diffusionseeder}. DiffusionSeeder detects obstacles using depth camera data rather than exact object poses. To ensure fairness, Euclidean signed distance field computation time is excluded from the results. As DiffusionSeeder inherently relies on camera input, it is not possible to provide it with exact obstacle information. This setup therefore represents the fairest possible comparison between the methods.
CAMPD achieves faster planning times than both cuRobo and DiffusionSeeder. Moreover, at higher batch sizes, CAMPD attains a higher success rate than cuRobo, as sampling more trajectories increases the likelihood of finding a better one. This improvement comes at only a minimal additional computational cost relative to cuRobo.
However, cuRobo consistently produces trajectories with the shortest motion time. \Cref{fig:example_mpinets} shows examples of tasks from the test set with a trajectory generated by CAMPD.

              \section{Conclusions} 
              \label{sec:conclusion}
              This work proposes Context-Aware Motion Planning Diffusion (CAMPD), a diffusion-based motion planning method capable of incorporating contextual information. CAMPD utilizes a diffusion model conditioned on contextual factors to estimate the added noise in trajectories. Experiments conducted in simulation demonstrate that CAMPD excels in generalizing to unseen environments, outperforming the existing method MPD in terms of success rate and solution optimality. Additionally, CAMPD showcases exceptional computational efficiency, making it suitable for online applications. Its ability to generate high-quality, executable trajectories directly on the robot further highlights its potential for practical deployment in complex motion planning tasks.
              While CAMPD shows promising results, several challenges remain. Training the model requires vast amounts of high-quality data, and the effectiveness of environment sampling strategies strongly influences performance. Moreover, CAMPD relies on structured representations such as spheres and cuboids, which can be difficult to replicate in real-world settings. Developing strategies to approximate generic objects with sets of geometric primitives from sensor data will be essential for real-world applicability. Finally, despite CAMPD’s fast runtime, its dependence on high-performance GPU hardware limits practicality in resource-constrained environments.
              Future work will therefore focus on reducing the dependency on large amounts of data, improving the model's generalization capabilities, reducing its computational complexity, and expanding its applicability to sensor-based environments.


              



\section*{Appendix}

\subsection{Model Parameters}
\label{sec:modelparams}
Diffusion steps were $T_{\text{train}} = 25$ with latent dimension $d_z = 64$. The U-Net had depth 4, 4 attention heads, and input dimension 64 per head. Training used batch size 128, learning rate $1\times10^{-4}$, and unconditional \text{probability $p_d = 0.33$}. Models ran on an NVIDIA\textsuperscript{\textcopyright} GeForce RTX\textsuperscript{\texttrademark} 4090 GPU. For DDPM, $T_{\text{inf}} = T_{\text{train}}$; for DPMSolver++, $T_{\text{inf}} = 3$. Final trajectories were smoothed using a Gaussian filter with standard deviation $\sigma = 2$ and window size 7.

\section*{Acknowledgments}
This work was supported by KU Leuven’s Special Research Fund (BOF), grant STG/25/004 (Advanced programming and control of smart and high-performance machines, vehicles, and robotic systems) and by Flanders Make through SBO project LearnOpTra (Learning meets optimization for robust and multimodal trajectory planning and control). Flanders Make is the Flemish strategic research
center for the manufacturing industry.


\bibliographystyle{IEEEtran}
\bibliography{root}  

\end{document}